\documentclass[lettersize, journal]{IEEEtran}
\usepackage{amsmath,amsfonts}
\usepackage{multirow, multicol}
\usepackage{algorithmic}
\usepackage{algorithm}
\usepackage{array}
\usepackage[caption=false,font=normalsize,labelfont=sf,textfont=sf]{subfig}
\usepackage{textcomp}
\usepackage{stfloats}
\usepackage{url}
\usepackage{verbatim}
\usepackage{graphicx}
\usepackage{xcolor, bm}
\usepackage{cite}
\newcommand{\revadd}[1]{{\color{black}#1}}
\begin{document}

\title{SparseLGS: Sparse View Language Embedded Gaussian Splatting}

\author{Jun~Hu,
        Zhang~Chen$^\dagger$,
        Zhong~Li,
        Yi~Xu,
        and~Juyong~Zhang$^\dagger$
\IEEEcompsocitemizethanks{\IEEEcompsocthanksitem J. Hu, and J. Zhang are with the School of Mathematical Science, University of Science and Technology of China.
Z. Chen is with the Meta, Z. Li is with the Apple Inc. and Y. Xu is with the Alpha Labs at Goertek.
}
\thanks{$^\dagger$Corresponding author. Email: \texttt{lansburyc@gmail.com, juyong@ustc.edu.cn}.}}



\maketitle

\begin{abstract}
Recently, several studies have combined Gaussian Splatting to obtain scene representations with language embeddings for open-vocabulary 3D scene understanding. While these methods perform well, they essentially require very dense multi-view inputs, limiting their applicability in real-world scenarios. In this work, we propose SparseLGS to address the challenge of 3D scene understanding with pose-free and sparse view input images. Our method leverages a learning-based dense stereo model to handle pose-free and sparse inputs, and a three-step region matching approach to address the multi-view semantic inconsistency problem, which is especially important for sparse inputs. Different from directly learning high-dimensional CLIP features, we extract low-dimensional information and build bijections to avoid excessive learning and storage costs. We introduce a reconstruction loss during semantic training to improve Gaussian positions and shapes. To the best of our knowledge, we are the first to address the 3D semantic field problem with sparse pose-free inputs. Experimental results show that SparseLGS achieves comparable quality when reconstructing semantic fields with fewer inputs (3-4 views) compared to previous SOTA methods with dense input. Besides, when using the same sparse input, SparseLGS leads significantly in quality and heavily improves the computation speed (5$\times$ speedup). Project page: {\tt\small \url{https://ustc3dv.github.io/SparseLGS}}
\end{abstract}


\section{Introduction}
\label{sec:intro}
\IEEEPARstart{3}{D} language field modeling is an important research problem in computer vision, offering extensive application prospects in fields such as autonomous driving, robotic manipulation~\cite{lerftogo2023, zheng2024gaussiangrasper}, and VR/AR. To obtain and enhance the quality of a 3D language field, high-precision 3D reconstruction is often necessary. Following the advent of NeRF~\cite{mildenhall2020nerf}, numerous works focusing on 3D semantic fields have emerged~\cite{semanticnerf:ICCV2021, liu2023semantic}. Initially, these semantic fields were more akin to rendering mask segmentation for each view, heavily reliant on semantic annotations of the data and lacking the capability for open language queries. To address these shortcomings, LERF~\cite{lerf2023} distills the required features from the language-image model CLIP and integrates them into NeRF. However, the bottlenecks of slow training and volumetric rendering in NeRF, as well as the quality limitations due to CLIP features being image-aligned rather than region or pixel-aligned, remain unresolved.

\begin{figure}
    \centering
    \includegraphics[width=\linewidth]{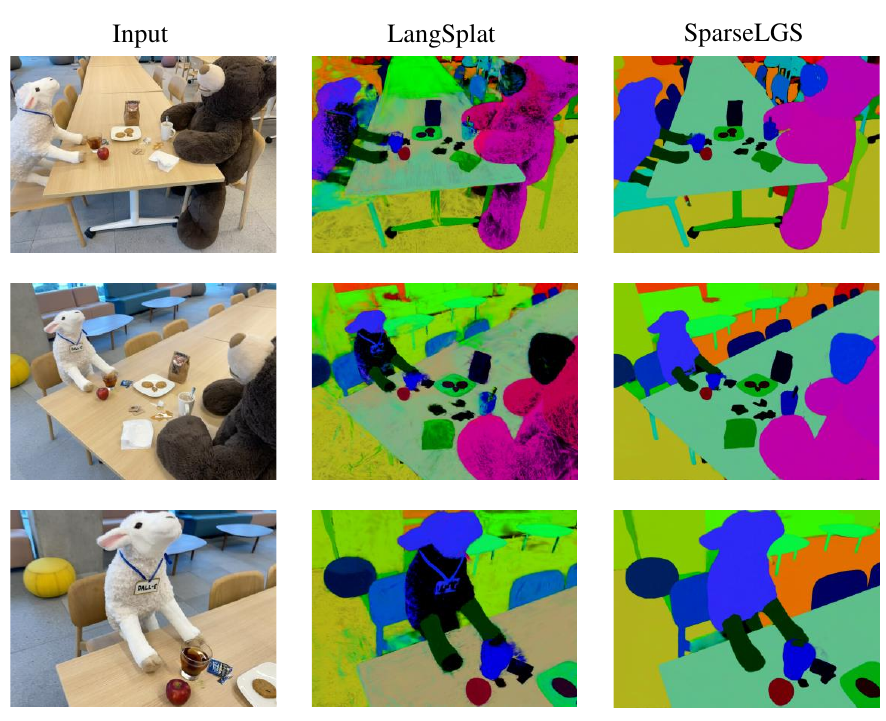}
    \caption{We present the semantic renderings from sparse, pose-free inputs using our method and LangSplat~\cite{qin2023langsplat}. Our method outperforms LangSplat in both multi-view consistency and rendering quality, producing more accurate and visually coherent results.}
    \label{fig:teaser}
\end{figure}

The recently proposed explicit 3D reconstruction method, 3D Gaussian Splatting~\cite{kerbl3Dgaussians}, offers fast training and real-time rendering, effectively addressing the speed issues associated with previous NeRF-based methods. Additionally, by using SAM~\cite{kirillov2023segany} for mask segmentation and integrating semantic models such as CLIP~\cite{cherti2023reproducible, openclip} or DINOv2~\cite{oquab2023dinov2}, it tackles the quality issues caused by unclear semantic boundaries. These methods~\cite{shi2023language, qin2023langsplat} optimize the semantics of Gaussians by downscaling the original CLIP features through techniques such as autoencoding and quantization with MLP. However, after obtaining the downscaled semantic features, they need to reconstruct the raw CLIP features. This restoration process can result in information loss, causing the resulted features to be inconsistent with the original features. In addition, like Gaussian Splatting, these methods usually require very dense input (usually more than 20 views) and highly accurate camera poses. The high input requirements and long training process make such methods difficult to apply in real-world scenarios. From a practical application standpoint, we prefer to use very sparse inputs (such as 3-4 images) to quickly obtain high-quality 3D language fields. This can significantly reduce the complexity of the data acquisition process and shorten training time, making it much more suitable for practical use.

In this paper, we propose Sparse View Language Embedded Gaussian Splatting (SparseLGS) to address the challenge of acquiring 3D language fields from sparse view inputs. To overcome the limitations of traditional off-the-shelf methods like COLMAP~\cite{schoenberger2016sfm, schoenberger2016mvs}, which often fail with extremely sparse views for point cloud reconstruction, we employ a learning-based dense stereo method, MASt3R~\cite{mast3r_arxiv24}, to estimate camera poses and generate the initial point cloud. Subsequently, we utilize SAM and CLIP to obtain object-level semantic results. In scenarios with dense view inputs, inconsistencies in multi-view semantics can be corrected because the abundance of views allows the accurate information to overshadow the few incorrect pieces. However, with sparse inputs (e.g., only 3-4 views), the incorrect results can distort the correct ones. The second column of Fig.~\ref{fig:teaser} shows the results of LangSplat~\cite{qin2023langsplat}, which serves as a typical example of how multi-view inconsistency leads to a degradation in rendering quality when using sparse view inputs. To address this issue, we adopt a three-step multi-view semantic alignment approach that utilizes techniques such as pixel matching and region fusion to achieve accurate alignment. To mitigate information loss during the reconstruction of the original features, we establish a bijection between the low-dimensional results and the original CLIP features. This allows us to use tile-based rendering to obtain the rendered semantic results and then utilize the bijection to restore the original CLIP features, thus enabling open-language queries.


Since the semantic masks provide regionalized information, with the interior of the same mask region being identical except for the boundary information, simply using semantic as ground truth does not provide sufficient geometric constraints. Therefore, we first train the Gaussian parameters using RGB images to initialize Gaussians. Subsequently, we incorporate a semantic loss to guide the training of the semantic field and fine-tune the Gaussian parameters.

To summarize, the contributions of this paper include:
\begin{itemize}
    \item We propose SparseLGS, which, to the best of our knowledge, is the first work to explore the reconstruction of 3D language fields from sparse pose-free view inputs.
    \item We propose ``three-step semantic multi-view matching" to resolve the inconsistencies in semantics and masks across input views. Additionally, we establish a bijection between the original CLIP features and the reduced-dimensional features to prevent degradation during the reconstruction of the original features.
    \item After optimizing the Gaussian parameters using RGB image supervision, we retain this supervision during the semantic field learning to better constrain the scene geometry. This strategy effectively enforces the 3D consistency of the learned semantic field under sparse input.
\end{itemize}

 
\section{Related Works}
\label{sec:related}
\subsection{3D Gaussian Splatting for 3D representation}
Unlike implicit reconstruction methods represented by NeRF~\cite{mildenhall2020nerf}, 3D Gaussian Splatting~\cite{kerbl3Dgaussians}, being an explicit model, is highly regarded for its ability to achieve real-time rendering while maintaining high-quality visual effects. Many approaches have combined 3D Gaussian Splatting to achieve improvements in both speed and quality. Some generalizable methods~\cite{liu2025mvsgaussian, chen2024mvsplat, charatan23pixelsplat} enhance the model's ability to generalize by extracting image features and integrating multi-view common information into the constructed neural network architecture. 3D surface reconstruction~\cite{Huang2DGS2024, guedon2023sugar}, generation~\cite{tang2023dreamgaussian, zhou2025dreamscene360} also use Gaussian Splatting and have achieved significant improvements in terms of visual effects and other related aspects. Some works~\cite{moreau2024human, xiang2024flashavatar, Gao2024PortraitGen} combine Gaussian Splatting to make their reconstruction of digital human and avatar much more efficient, with higher quality and better editability. Different from the applications mentioned in the above works, we aim to leverage language-embedded Gaussians to better construct a 3D language field to support open-vocabulary queries.

\subsection{Sparse View 3D Reconstruction}
3D reconstruction tasks often require dense views and precise camera poses for supervision. Due to the difficulty of meeting these needs, a series of works circumvent the requirements of dense input views. BARF~\cite{lin2021barf} and NeRF$--$~\cite{wang2021nerfmm} jointly optimize the radiance field and camera parameters with initial noise. GARF~\cite{chng2022garf} proposes a matching method and uses a different activation function to ease the pose estimation. Simple-RF~\cite{somraj2024simplerf} chooses to reduce the fitting and expressive capabilities of NeRF-like models, and HG3-NeRF~\cite{gao2024hg3nerf} uses CLIP feature to assist in this coarse-to-fine reconstruction process. SPARF~\cite{sparf2023} uses virtual views and pixel matching, designing two related losses to help optimize camera poses. These works are all related to NeRF. As Gaussian Splatting becomes increasingly popular, a significant amount of work has emerged focusing on sparse reconstruction based on it because of its effectiveness and high quality. CoR-GS~\cite{zhang2024corgs} simultaneously trains two Gaussian fields and optimizes based on the inconsistencies between these two fields. DNGaussian~\cite{li2024dngaussian} and FSGS~\cite{zhu2023FSGS} emphasize depth information and focus on optimizing the Gaussian distribution using both global and local information. These methods focus on learning RGB and do not address the issue of 3D semantic field reconstruction. Therefore, we utilize the learning-based MASt3R to provide excellent camera poses and point clouds to address the challenge of sparse reconstruction of 3D semantic fields.

\subsection{3D Language Fields}
After making significant progress in 2D semantics in computer vision, researchers have begun to venture into the more challenging domain of 3D semantic fields. Semantic-NeRF~\cite{semanticnerf:ICCV2021} easily combine semantic masks with NeRF to get a 3D segmentation field. GSNeRF, Semantic-Ray, RT-GS2~\cite{liu2023semantic, Chou2024gsnerf, jurca2024rtgs2} have developed different network architectures and pipelines, resulting in the training of generalizable scene segmentation models. \revadd{Large Spatial Model (LSM)~\cite{fan2024largespatialmodel} further studies feed-forward semantic 3D prediction from unposed images and predicts geometry, appearance, and semantic Gaussian fields.} The aforementioned methods are capable of achieving 3D semantic segmentation, but they can not perform text-to-image content queries. Subsequently, many methods have been developed to get open-ended 3D language fields based on CLIP features. Feature 3DGS~\cite{zhou2023feature} uses SAM feature to get 3D segmentation field and uses CLIP to enable text-to-specific-object queries. CLIP-GS~\cite{liao2024clip} focuses on real-time 3D semantic understanding of videos and employs a codebook for dimensionality reduction. LEGaussians~\cite{shi2023language} combines both DINOv2 and CLIP and uses MLP with softmax to obtain the semantic feature. LangSplat~\cite{qin2023langsplat} uses autoencoder to reduce the CLIP feature's dimensionality and restore the encoded feature and FastLGS~\cite{ji2024fastlgs} uses feature grid to distinguish and bind the mapping from high-dimensional to low-dimensional features. Gaussian Grouping~\cite{gaussian_grouping} uses SAM along with a compact identity encoding to get 3D understanding. OpenGaussian~\cite{wu2024opengaussian} constrains spatial semantics in 3D and uses a coarse-to-fine codebook for object semantic differentiation. Unlike the aforementioned methods, we focus on how to efficiently obtain high-quality 3D language fields from pose-free sparse inputs to support open-vocabulary queries.

\section{Method}
\begin{figure*}[htbp]
    \centering
    \vspace*{-10pt}
    \includegraphics[width=\linewidth]{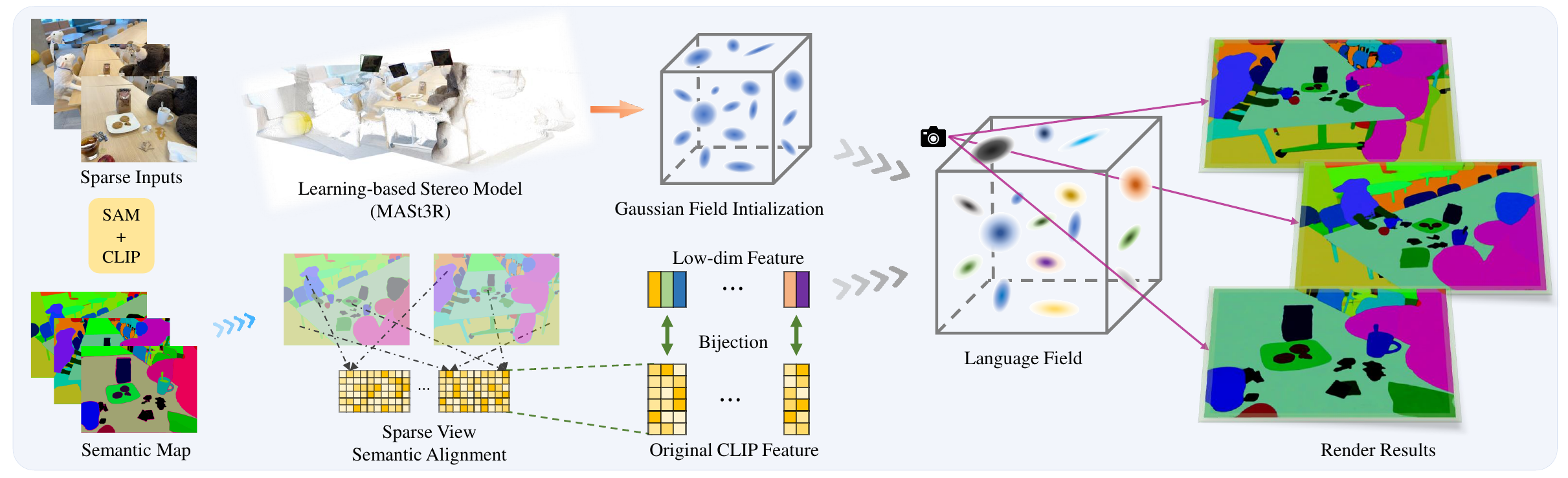}
    \vspace*{-17pt}
    \caption{Our approach \textbf{SparseLGS} is capable of generating high-quality language fields from pose-free sparse view inputs in just a few minutes. We first leverage SAM and CLIP to obtain object-wise semantic maps, then use a learning-based stereo model to derive camera poses and point clouds from sparse inputs. To address semantic inconsistencies across views, we employ a three-step multi-view semantic alignment strategy. To better integrate semantics with Gaussian Splatting, we establish a bijection between the original CLIP features and their dimensionality-reduced counterparts. During training, we incorporate RGB supervision to enhance the 3D consistency of our learned language field.}
    \vspace*{-5pt}
    \label{fig:pipline}
\end{figure*}

The whole pipeline is illustrated in Fig.~\ref{fig:pipline}. We provide a brief introduction to Gaussian Splatting and describe how to obtain object-wise semantic features for semantic field training in Section~\ref{sec:preli}. In Section~\ref{sec:camera}, we introduce the multi-view stereo models to accurately estimate camera poses and generate initial point clouds. We address the issue of multi-view inconsistencies under sparse inputs in Section~\ref{sec:align}. Finally, we elaborated on our two-stage training ideas and specific practices in Section~\ref{sec:train}.

\subsection{Preliminary}
\label{sec:preli}
Gaussian Splatting~\cite{kerbl3Dgaussians} is an explicit 3D scene representation approach, where the entire scene is explicitly modeled as a series of anisotropic 3D Gaussians. Using these 3D Gaussian primitives along with the camera's intrinsic and extrinsic parameters, the color $\mathbf{C}$ for each pixel can be computed.

Specifically, each 3D gaussian can be parameterized by a mean vector $\bm{\mu} \in \mathbb{R}^3$ and a covariance matrix $\bm{\Sigma} \in \mathbb{R}^{3\times 3}$:
\begin{equation}
    G(\mathbf{x}) = e^{-\frac{1}{2}(\mathbf{x} - \bm{\mu})\bm{\Sigma}^{-1}(\mathbf{x} - \bm{\mu})}.
\end{equation}

To ensure that $\bm{\Sigma}$ is positive semi-definite, we represent it using a scaling matrix $\mathbf{S}$ and a rotation matrix $\mathbf{R}$ such that $\bm{\Sigma} = \mathbf{R}\mathbf{S}\mathbf{S}^{T}\mathbf{R}^{T}$. Finally, the 3D Gaussians are efficiently rendered onto a 2D image plane using tile-based rasterization. The alpha blending process proceeds as follows:
\begin{equation}
    \mathbf{C} = \sum_{i \in \mathcal{N}} \mathbf{c}_i \alpha_i \prod_{j=1}^{i-1}(1-\alpha_j),
\end{equation}
where $\mathbf{c}_i$ represents the color of each Gaussian, $\mathcal{N}$ denotes the collection of Gaussians that the ray intersects, and $\alpha_i = o_i G_i^{2D}$, where $\alpha_i$ is composed of the opacity $o_i$ of the $i$-th Gaussian and the 2D projection $G_i^{2D}$ of the $i$-th Gaussian.

To achieve semantic Gaussian Splatting, each Gaussian is additionally assigned a semantic feature $\mathbf{f}_i$. Therefore, similar to the previous rendering process, we can also obtain the rendered semantic features through alpha blending as follows:
\begin{equation}
    \mathbf{F}_{\textrm{lan}} = \sum_{i \in \mathcal{N}} \mathbf{f}_i \alpha_i \prod_{j=1}^{i-1}(1-\alpha_j).
\end{equation}

To optimize these $\mathbf{f}_i$ with object-wise semantic features, we use the SAM~\cite{kirillov2023segany} model to get the image's object segmentation and the CLIP~\cite{cherti2023reproducible, openclip} model to obtain the semantic information of each object region, instead of relying on the unclear patch-wise semantic features from DINOv2~\cite{oquab2023dinov2}.

\subsection{Camera Pose and Point Cloud Estimation}
\label{sec:camera}
First, we need to estimate the camera pose and the initial point cloud from sparse inputs to train these Gaussians. Current methods typically rely on Structure from Motion~\cite{schoenberger2016sfm} (SfM) and Multi-View Stereo~\cite{schoenberger2016mvs} (MVS) to precompute camera poses and sparse point clouds from dense inputs. While this approach is effective for 3D reconstruction with dense views, it often fails to estimate correct poses when the input views are sparse and exhibit significant variations in viewpoints (e.g., with three views and camera angle differences exceeding 90 degrees). Therefore, directly applying methods similar to COLMAP may not yield accurate initializations.

Recently, new models such as DUSt3R~\cite{dust3r_arxiv23, dust3r_cvpr24} and MASt3R~\cite{mast3r_arxiv24} have integrated the SfM and MVS processes into a single pipeline, enabling end-to-end reconstruction of camera poses and dense point clouds from pose-free sparse view inputs. By replacing the COLMAP process with these methods, a robust initialization is provided, significantly improving the issue of poor sparse reconstruction quality caused by limited input constraints. This forms a solid foundation for enhancing the quality of the 3D semantic field.

\subsection{Sparse View Semantic Alignment}
\label{sec:align}
We first introduce our inputs and corresponding notations. Given a set of input images $\{I_{t}| t = 1, \dots, T\}$, for each image $I_t$, we can get whole segmentation masks $\{M_{t,j}^{s}|j=1,\dots,m_t \textrm{ and } s = 1,2,3\}$ for three different granularities (whole, subpart, part) and compute the corresponding CLIP features $\{\mathbf{L}_{t,j}^{s}|j=1,\dots,m_t \textrm{ and } s = 1,2,3\}$. 

Now that we have obtained camera pose, initial point cloud, and semantic map through the previous data preprocessing steps, we can begin training the 3D semantic field. However, under the setting of sparse view inputs, a significant challenge remains. Specifically, for the same object viewed from different perspectives, ensuring 3D consistency in semantics becomes difficult due to factors such as view direction, cluttered backgrounds, and occlusions. When dense input views are available, slight inconsistencies can be averaged out through a sufficiently large number of training samples. However, as the number of views decreases, semantic inconsistencies between different views become more pronounced and severe. These inconsistencies degrade the effectiveness of the trained 3D semantic field and lead to a reduction in the accuracy of text queries.

To mitigate the impact of sparse view semantic inconsistency, we propose a semantic alignment method consisting of three parts: RoMa-based pixel matching, inconsistent mask fusion, and reprojection matching fine-tuning. \revadd{These stages target different ambiguity sources rather than repeating the same matching operation. Pixel matching first provides cross-view correspondence candidates, mask fusion handles inconsistent SAM mask granularity, and reprojection matching uses the estimated 3D geometry to filter correspondences that remain ambiguous in 2D.} Doing so, we can address not only semantic inconsistencies (step 1, 3) but also discrepancies in mask segmentation (step 2).

\paragraph{Step1: RoMa-based pixel matching.} First, we use RoMa~\cite{edstedt2024roma} to complete the matching between different semantic masks. For images $I_i$ and $I_j$, assume that the mask area in $I_i$ is $M_i=\{p_k\}_{k=1}^N$. Each pixel $p_k$ in $M_i$ can find a corresponding match $q_k$ in $I_j$. These $\{q_k\}_{k=1}^N$ will each belong to different semantic masks in $I_j$. The SAM mask $M_j$ with the highest number of matching points is the semantic mask that $M_i$ matches in $I_j$. Then, using the matching area ratio $S^{\textrm{area}}_{ij}$ between $M_i$ and $M_j$, along with the cosine distance $S^{\textrm{lang}}_{ij}$ between the corresponding features $\mathbf{L}_i$ and $\mathbf{L}_j$ (as defined in Equation~\ref{equ:dis}), we can evaluate the alignment consistency between masks.
\begin{equation}
    S^{\textrm{area}}_{ij} = \frac{\#\{p_k \in M_j\}_{k=1}^N}{\#M_j};~ S^{\textrm{lang}}_{ij} = \frac{\mathbf{L}_i \cdot \mathbf{L}_j}{\|\mathbf{L}_i\|\|\mathbf{L}_j\|}.
    \label{equ:dis}
\end{equation}

The truly matched SAM mask pairs $(M_i, M_j)$ can be selected when $S_{ij}^{\textrm{match}} > \tau_1$, where $\tau_1$ controls the confidence level of the filtering process. Here, $S_{ij}^{\textrm{match}}$ is defined in Equation~\ref{equ:match}, where $\lambda$ represents the weight of $S^{\textrm{lang}}_{ij}$
\begin{equation}
    S_{ij}^{\textrm{match}} = \lambda S^{\textrm{lang}}_{ij} + (1 - \lambda) S^{\textrm{area}}_{ij}.
    \label{equ:match}
\end{equation}

\paragraph{Step2: Inconsistent mask fusion.} After matching, the semantic inconsistency problem is resolved. However, inconsistencies in SAM segmentation across different views persist. For example, two regions may appear within the same mask in $I_i$ but belong to different masks in $I_j$. 
    
For coarser segmentation, we aim for each mask to represent a complete object. Based on the previously matched mask pairs, if multiple masks in $I_i$ correspond to the same mask $M_j$ in $I_j$, and these pairs meet the screening criteria, we merge the mask regions in $I_i$ and assign them the semantics of $M_j$. For finer segmentation, however, we avoid mask fusion to ensure that the same object can be divided into smaller, more detailed segments. Specifically, we adopt three granularities of semantic segmentation. At the finest level, we skip mask fusion to preserve the separation of finely segmented regions and avoid merging during fusion. This ensures the independence of segmentation results across different granularities.

\paragraph{Step3: Reprojection matching fine-tuning.}
After the previous two steps, the issue of semantic inconsistency across sparse views is largely alleviated. However, RoMa may struggle to accurately match points that are spatially close but observed from significantly different viewing angles. To address this, we use the pixel's corresponding 3D position to assist in refining the matches.

Specifically, for a SAM mask $M_i \in I_i$, each pixel in $M_i$ can be back-projected into 3D space and then reprojected onto another view, such as $I_j$. Similar to Step 1, the corresponding mask $M_j$ in $I_j$ can be identified. For $M_j$, we can likewise find the corresponding mask $\hat{M}_i$ in $I_i$ through back-projection. The bilateral matching results could be calculated respectively using $S_{ij}^{\textrm{match}}$ as in step 1. The correct SAM mask pairs $(M_i, M_j)$ are retained if $M_i = \hat{M}_i$ and $S_{ij}^{\textrm{match}} > \tau_2$. 

\subsection{Training Sparse View 3D Language Fields}
\label{sec:train}
In previous dense-input 3D language field representation methods, RGB supervision is abandoned when training semantic features. However, if we rely solely on semantic loss to train the Gaussians in the sparse input setting, they tend to become excessively elongated or overly large, failing to capture the correct geometric distribution of the scene. This is entirely due to the semantic map providing minimal information that is highly regionalized, with almost no additional information contained within the interior of each region. This leads to the Gaussian shape being able to grow indiscriminately and not being well controlled. In contrast, the RGB image contains richer information and could provide much stronger geometric constraints. Therefore, we first train the Gaussians without semantic constraints, which serves as a robust initialization for modeling the 3D semantic field. Additionally, during the initial training of the Gaussians, we incorporate camera pose optimization to correct slight errors in the estimated camera poses. The training process is as follows:
\begin{equation}
    \mathcal{L}_{\textrm{img}} = \lambda_1\mathcal{L}_1 + (1-\lambda_1)\mathcal{L}_{\textrm{SSIM}}.
\end{equation}

\paragraph{Bijection.} During semantic training, if we directly combine hundreds of thousands of Gaussians with CLIP features, it results in unacceptable storage overhead and inefficient rendering and training. To address this, we need to reduce the dimensionality of the raw semantic features. Current methods~\cite{qin2023langsplat, shi2023language} typically rely on training an autoencoder for dimensionality reduction or use quantization and MLPs. However, both approaches suffer from the issue that the \textcolor{black}{inverse mapping of low-dim features} can not align well with the original CLIP features.

Our solution is to perform dimensionality reduction on the original features using techniques such as PCA, MLP, or one-dimensional convolution. A single scene typically requires only a few hundred or at most a thousand semantic labels to represent all objects. During dimensionality reduction, we bind each low-dim feature to its corresponding high-dim semantic feature. In evaluation, when a low-dim feature map is rendered, we identify the most similar feature from these low-dim features and retrieve its associated high-dim semantic feature, thereby completing the inverse mapping. This approach minimizes errors caused by reconstructing the original features. We denote the low-dim semantic features in $I_i$ as $\{\hat{\mathbf{L}}_{i,j}^{s}|j=1,\dots,m_i \textrm{ and } s = 1,2,3\}$

\paragraph{RGB-assisted semantic training.} Subsequently, during the training of semantic features, to ensure that the Gaussian properties change only slightly, apart from the semantic properties, and to provide some geometric constraints, the loss function for training the semantic Gaussians combines the image loss with the semantic loss. Let $\mathbf{F}_{i,j}^s$ denote the corresponding rendered semantic features for segmentation level $s$, and the total loss function can be expressed as:
\begin{align}
    \mathcal{L} = \lambda_2 \mathcal{L}_{\textrm{img}} + (1-\lambda_2) \mathcal{L}_{\textrm{sem}},
\end{align}
where $\mathcal{L}_{\textrm{sem}} = \sum_{i=1}^T\sum_{j=1}^{m_i}\|\mathbf{F}_{i,j}^s - \hat{\mathbf{L}}_{i,j}^s\|$.





\section{Implementation Details}
\subsection{About the Experimental Setup}
We implement our framework using PyTorch\cite{paszke2019pytorch} and incorporate the modified CUDA kernels from 3D Gaussian Splatting to enable the rendering of semantic features. During the initialization training of the Gaussian parameters, we incorporate the prune and densify process. This process is not performed during the semantic training. We set $\lambda_1 = 0.8$ and $\lambda_2 = 0.3$ in our training stage and set $\tau_1 = 0.5, \tau_2 = 0.5$, $\lambda = 0.3$ in sparse view semantic alignment stage. These parameters could be fine-tuned for different scenes, which could make the visual quality better. Due to the sparsity of our inputs, We require approximately 30 seconds to estimate camera poses and point clouds, around 4 minutes to obtain the semantic segmentation, and about 30 seconds to complete multi-view semantic alignment. Our model takes approximately 3 minutes to complete the semantic training on one RTX3090 GPU. We use the Adam~\cite{KingBa15} optimizer for training, with a learning rate set to $10^{-8}$ for semantic features. Due to the good initialization, each of the three granularity levels of the semantic Gaussian fields is trained for only 1000 iterations.

\subsection{Relevancy Score of Open-vocabulary Queries}
Inspired by LERF~\cite{lerf2023}, we fisrt compute the CLIP embedding of the text query $\phi_{\textrm{quer}}$ and a set of canonical phrases $\phi_{\textrm{canon}}^i$. Then compute cos similarity of the rendered semantics and each $\phi_{\textrm{canon}}^i$. Finally, compute the pairwise softmax between the semantic map $\phi_{\textrm{sem}}$ and the text embedding. We could get the relevancy score below:
\begin{equation}
    \min_i \frac{\exp(\phi_{\textrm{sem}} \cdot \phi_{\textrm{quer}})}{\exp(\phi_{\textrm{sem}} \cdot \phi_{\textrm{canon}}^i) + \exp(\phi_{\textrm{sem}} \cdot \phi_{\textrm{quer}})},
\end{equation}
Where canonical phrases ``object", ``things", ``stuff", ``texture" are used for all experiments. This score quantifies the correlation between the rendered semantics and the text query's CLIP embedding.

Note that we train three different granularity levels of semantic fields. For each query, we set the most relevant one (with highest relevancy score) as the result relevancy map.

\subsection{The Evalution Metrics}
We use Mean Intersection over Union (mIoU) and Mean Accuracy (mAcc) to measure the performance of each method in open-vocabulary semantic mask query tasks and semantic localization tasks. For each specific object query, we search within each granularity using different strategies to decide the best semantic level.

\paragraph{LERF Dataset.}
We firstly use mean convolution filter to mitigate the impact of outliers in each relevancy map. For 3D object localization tasks, we select the one with the highest relevancy score and its corresponding position as our predicted localization. For 3D semantic segmentation tasks, we select the area where the relevancy score surpasses our preset threshold (the default value is 0.6) as our predicted semantic mask. Our numerical estimations and visualization results on the LERF dataset are based on adaptations of the evaluation code from LangSplat.

\paragraph{3D-OVS Dataset.}
For 3D-OVS dataset, select the region with a relevancy score exceeding the threshold $\tau$ as the corresponding mask. The threshold $\tau$ can be fine-tuned for various datasets to achieve more optimized masks. Here, the most relevant mask across three distinct granularity levels is determined based on both the area and the average relevancy score within the mask. Specifically, we calculate the area and the average score of each level's mask, the one with an area greater than 2000 that also achieves the highest score will be selected as our final output. The default setting for $\tau$ is 0.8.
\section{Experiments}

\begin{table*}[htbp]
\centering
\caption{IoU scores, localization accuracy, and time comparisons on the LERF dataset. DS. denotes using a dense stereo model, DC. refers to using COLMAP with dense views. Tr.T. represents the training time, and T.T. represents the total time. Note that COLMAP fails to obtain camera poses and point clouds from sparse inputs. The \colorbox[HTML]{FD6864}{\textbf{best}} and \colorbox[HTML]{FFCE93}{\textbf{second best}} scores are highlighted in different colors for clarity.}
\resizebox{\linewidth}{!}{
\begin{tabular}{c|cccc|c|cccc|c|cc}
    & \multicolumn{5}{c|}{mIOU $\uparrow$} & \multicolumn{5}{c|}{mACC $\uparrow$} & 
\multirow{2}{*}{Tr.T $\downarrow$} & \multirow{2}{*}{T.T $\downarrow$}\\
    & ramen & figur. & tea. & kitchen & mean & ramen & figur. & tea. & kitchen & mean \\
    \hline
    DC. + LangSplat~\cite{qin2023langsplat} & \colorbox[HTML]{FFCE93}{0.291} & \colorbox[HTML]{FFCE93}{0.347} & 0.630 & \colorbox[HTML]{FD6864}{0.551} & \colorbox[HTML]{FFCE93}{0.455} & 0.385 & 0.571 & 0.842 & \colorbox[HTML]{FFCE93}{0.667} & 0.616 & 20 min & 70 min \\
    
    DC. + LEGaussian~\cite{shi2023language} & 0.222 & 0.249 & 0.328 & 0.213 & 0.253 & \colorbox[HTML]{FFCE93}{0.538} & 0.411 & 0.842 & \colorbox[HTML]{FFCE93}{0.667} & 0.614 & 25 min & 30 min\\
    
    DS. + LangSplat~\cite{qin2023langsplat} & 0.265 & 0.328 & \colorbox[HTML]{FFCE93}{0.634} & 0.478 & 0.426 & 0.462 & \colorbox[HTML]{FFCE93}{0.607} & \colorbox[HTML]{FD6864}{0.921} & 0.556 & \colorbox[HTML]{FFCE93}{0.636} & \colorbox[HTML]{FD6864}{40s} & \colorbox[HTML]{FD6864}{6 min}\\
    
    SparseLGS & \colorbox[HTML]{FD6864}{0.402} & \colorbox[HTML]{FD6864}{0.487} & \colorbox[HTML]{FD6864}{0.713} & \colorbox[HTML]{FD6864}{0.551} & \colorbox[HTML]{FD6864}{0.538} & \colorbox[HTML]{FD6864}{0.692} & \colorbox[HTML]{FD6864}{0.732} & \colorbox[HTML]{FFCE93}{0.868} & \colorbox[HTML]{FD6864}{0.889} & \colorbox[HTML]{FD6864}{0.770} & \colorbox[HTML]{FD6864}{40s} & \colorbox[HTML]{FD6864}{6 min}\\
\end{tabular}
}
\label{tab:lerfovs}
\end{table*}

\begin{table*}[htp]
\centering
\caption{We present IoU scores and time comparisons on the 3D-OVS dataset. S.C. denotes COLMAP with sparse views, and D.S. refers to using a dense stereo model. The \colorbox[HTML]{FD6864}{\textbf{best}} and \colorbox[HTML]{FFCE93}{\textbf{second best}} scores are highlighted in different colors for emphasis.}
\begin{tabular}{c|ccccc|c|cc}
    & \multicolumn{6}{c|}{mIOU $\uparrow$} & \multirow{2}{*}{Tr.T $\downarrow$} & \multirow{2}{*}{T.T $\downarrow$}\\
    & bed & sofa & bench & room & desk & mean\\
    \hline
    SC. + LangSplat~\cite{qin2023langsplat} & 0.558 & 0.789 & 0.786 & 0.004 & \colorbox[HTML]{FFCE93}{0.868} & 0.601 & 27 min & 100 min\\
    SC. + LEGaussian~\cite{shi2023language} & 0.418 & 0.456 & 0.328 & \colorbox[HTML]{FFCE93}{0.386} & 0.400 & 0.397 & 35 min & 40 min\\
    DS. + LangSplat~\cite{qin2023langsplat} & \colorbox[HTML]{FFCE93}{0.628} & \colorbox[HTML]{FD6864}{0.866} & \colorbox[HTML]{FFCE93}{0.841} & 0.004 & \colorbox[HTML]{FD6864}{0.884} & \colorbox[HTML]{FFCE93}{0.645} & \colorbox[HTML]{FD6864}{1 min} & \colorbox[HTML]{FD6864}{8 min}\\
    SparseLGS & \colorbox[HTML]{FD6864}{0.911} & \colorbox[HTML]{FFCE93}{0.831} & \colorbox[HTML]{FD6864}{0.903} & \colorbox[HTML]{FD6864}{0.730} & 0.843 & \colorbox[HTML]{FD6864}{0.844} & \colorbox[HTML]{FD6864}{1 min} & \colorbox[HTML]{FD6864}{8 min}\\
\end{tabular}
\label{tab:3dovs}
\end{table*}

\subsection{Datasets and Baseline}
We conduct experiments on two widely-used datasets: 3D-OVS~\cite{3dovs} and LERF~\cite{lerf2023}. The LERF dataset consists of 13 scenes containing a mixture of in-the-wild and posed long-tail scenes. It features complex and varied scenes with a wide range of objects, showcasing the method's capability to handle real-world data. Inspired by LERF and LangSplat~\cite{qin2023langsplat}, we use the mIoU metric to evaluate the quality of the predicted mask for open-vocabulary queries and mACC to assess accuracy in the object localization task. The 3D-OVS dataset comprises distinct scenes with a set of long-tail objects situated in various poses and backgrounds, making it well-suited for evaluating the quality of object masks under open-vocabulary tasks. Therefore, we use mIoU as the evaluation metric for the 3D-OVS dataset.

We compare our SparseLGS with recent SOTA language-embedded Gaussian splatting methods such as LangSplat~\cite{qin2023langsplat} and LEGaussian~\cite{shi2023language}. For the LERF dataset, using COLMAP does not yield camera poses due to the very sparse views and the complexity of the scene. Therefore, when conducting experiments on the LERF dataset, we use all images to obtain camera poses and initial point clouds. Additionally, since there is no existing work specifically addressing sparse reconstruction of 3D language fields, \textcolor{black}{we create a simple combination of MASt3R $+$ LangSplat as our baseline and comparative method.}
\revadd{We further include Large Spatial Model (LSM)~\cite{fan2024largespatialmodel} as a representative feed-forward baseline. Because LSM learns semantic fields from LSeg feature supervision on ScanNet-style data, we report comparisons under the ScanNet sparse-view setting and the LERF Teatime transfer setting. This helps separate feed-forward reconstruction capability from semantic-source and dataset-distribution differences.}

\begin{figure*}[htbp]
    \centering
    \includegraphics[width=\linewidth]{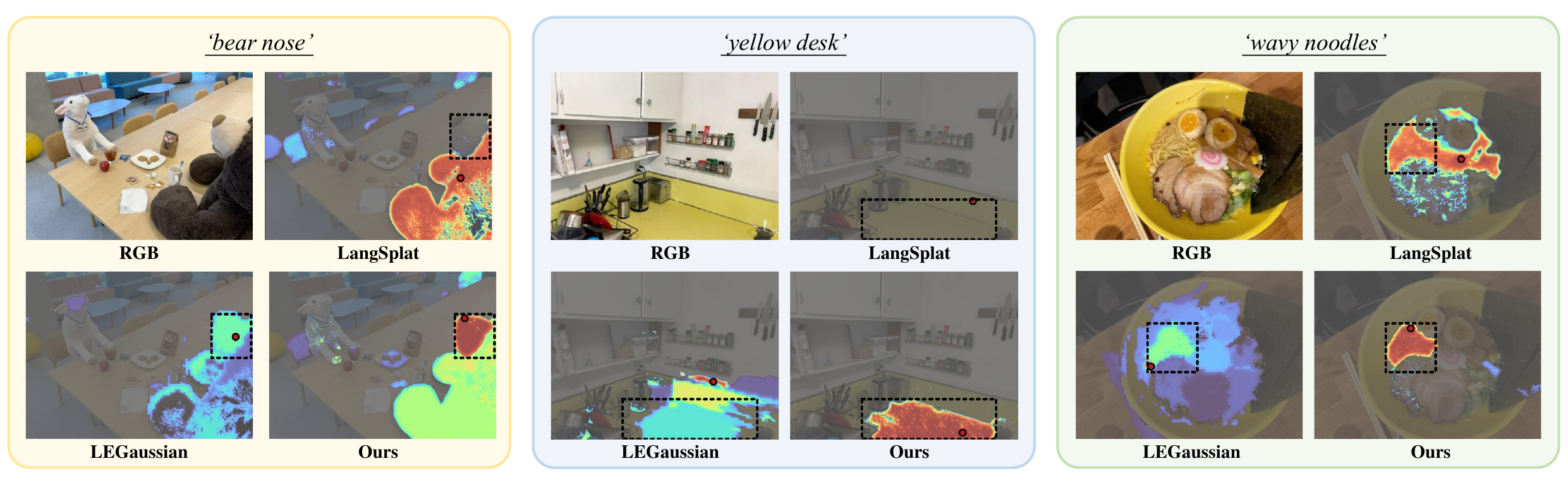}
    \caption{Open-vocabulary 3D object localization experiments on the LERF datasets. The black dashed box represents the GT bounding box of the query object, while the red dots indicate the predicted locations of the query objects by each method.}
    \label{fig:lerf_loc}
\end{figure*}

\begin{figure*}[htbp]
    \centering
    \includegraphics[width=\linewidth]{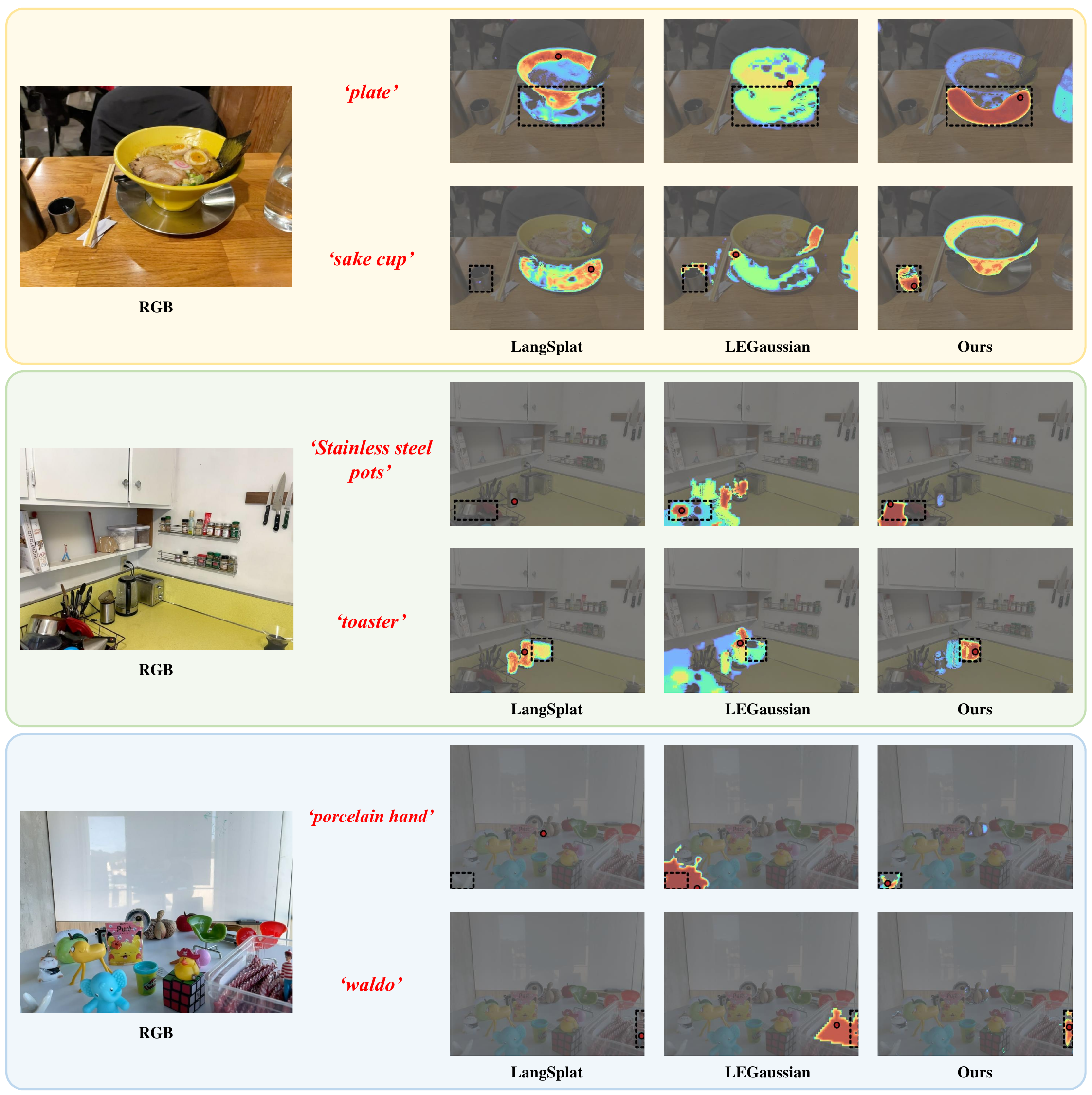}
    \caption{More open-vocabulary 3D object localization experiments on the LERF datasets. The black dashed box represents the GT bounding box of the query object, and the red dots indicate the predicted locations of the query objects by each method.}
    \vspace*{-7pt}
    \label{fig:supp_lerf_loc}
\end{figure*}

\begin{figure*}[htbp]
    \centering
    \includegraphics[width=\linewidth]{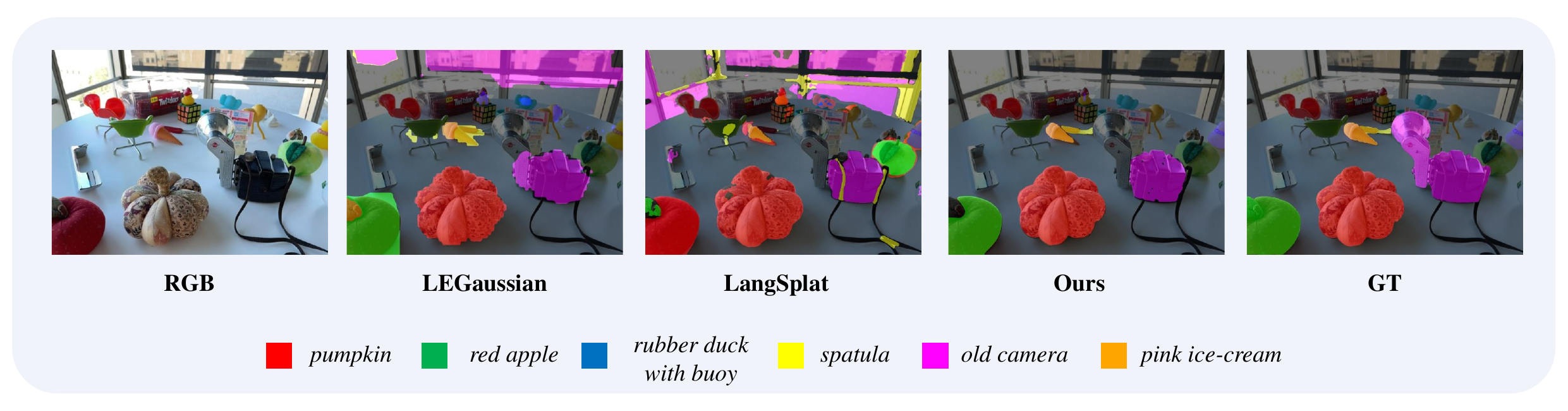}
    \vspace{-15pt}
    \caption{Open-vocabulary 3D semantic segmentation on the LERF dataset.}
    \label{fig:lerf_ex}
\end{figure*}

\subsection{Evaluation Results}
\subsubsection{LERF Dataset}
Table~\ref{tab:lerfovs} shows the quantitative Results of our method in object localization and semantic segmentation compared to other methods. We use four views for all the experiments on LERF dataset. ``DC." denotes using dense inputs (e.g., all images) to obtain camera poses and point clouds through COLMAP. The reason we did this is that these methods use COLMAP as initialization for dense inputs, but COLMAP cannot provide camera poses for very sparse inputs (3-4 views) with significant view changes. Therefore, we relaxed the conditions for initialization in these methods. As shown in Table~\ref{tab:lerfovs}, even with more information given to these methods, our method still achieves much better results in object localization and semantic segmentation tasks. Moreover, due to our use of multi-view stereo for obtaining great initialization, we only need to train our model for 1k iterations, which makes us much faster than other methods in terms of training time (Tr.T.) and total time (almost five times faster). `total time' here refers to the sum of data preprocessing time and training time, denoted by T.T.

\begin{figure*}[htbp]
    \centering
    \includegraphics[width=\linewidth]{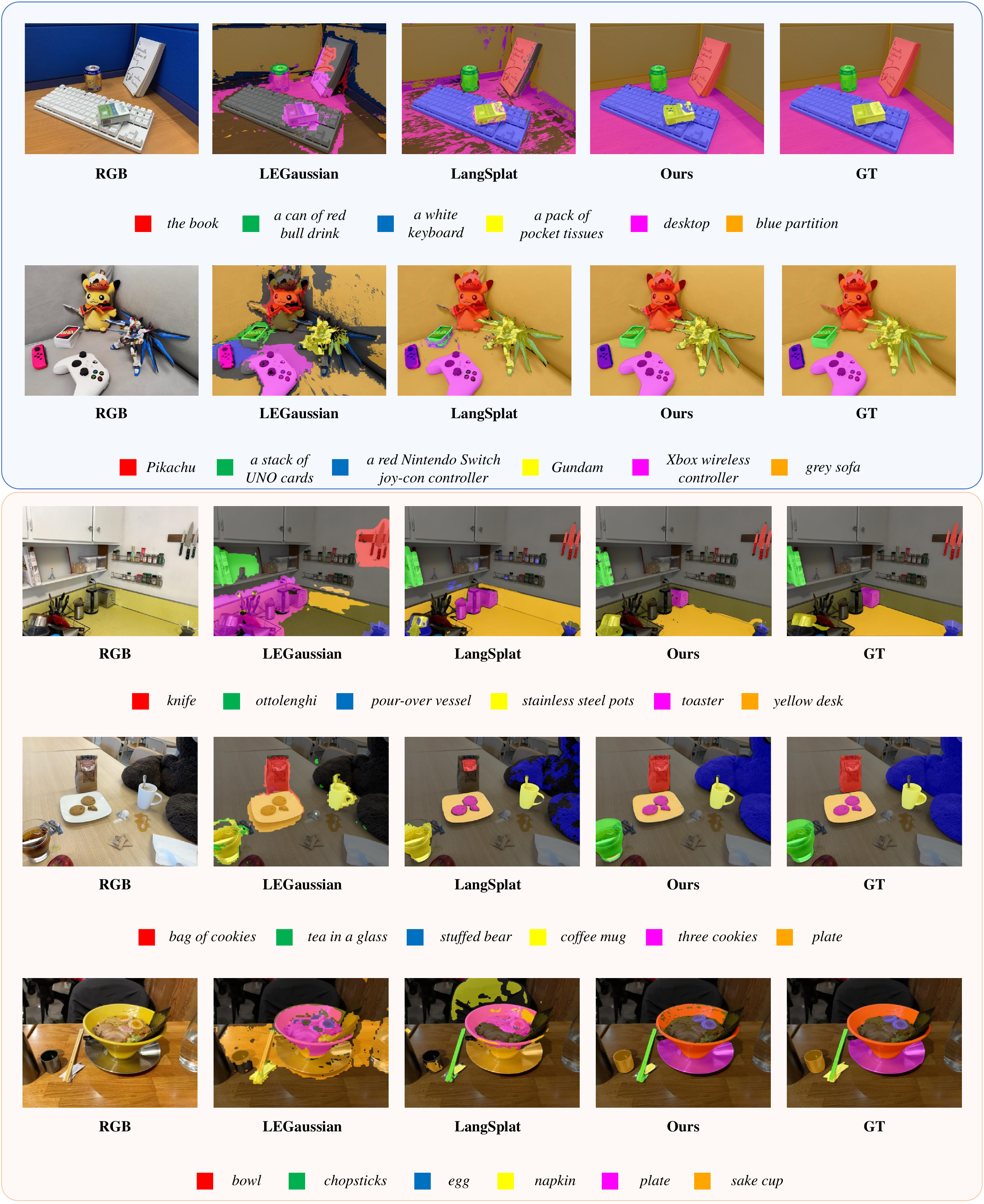}
    \vspace{-15pt}
    \caption{More open-vocabulary 3D semantic segmentation experiments on the LERF datasets and 3D-OVS datasets.}
    \label{fig:3dovs_supp}
\end{figure*}

Fig.~\ref{fig:lerf_loc} and Fig.~\ref{fig:supp_lerf_loc} display the qualitative comparisons of 3D object localization task of each method. It can be observed that under sparse inputs, we are able to more accurately locate the positions of objects, and the relevance heat map also indicates that our predicted regions are highly concentrated. Fig.~\ref{fig:lerf_ex} and the brown box of Fig.~\ref{fig:3dovs_supp} show the comparisons on 3D semantic segmentation tasks with open-vocabulary queries. It is evident that the semantic mask regions we obtained closely match the Ground Truth (GT) and essentially do not query any irrelevant semantic parts.

\begin{figure*}[htbp]
    \centering
    \includegraphics[width=\linewidth]{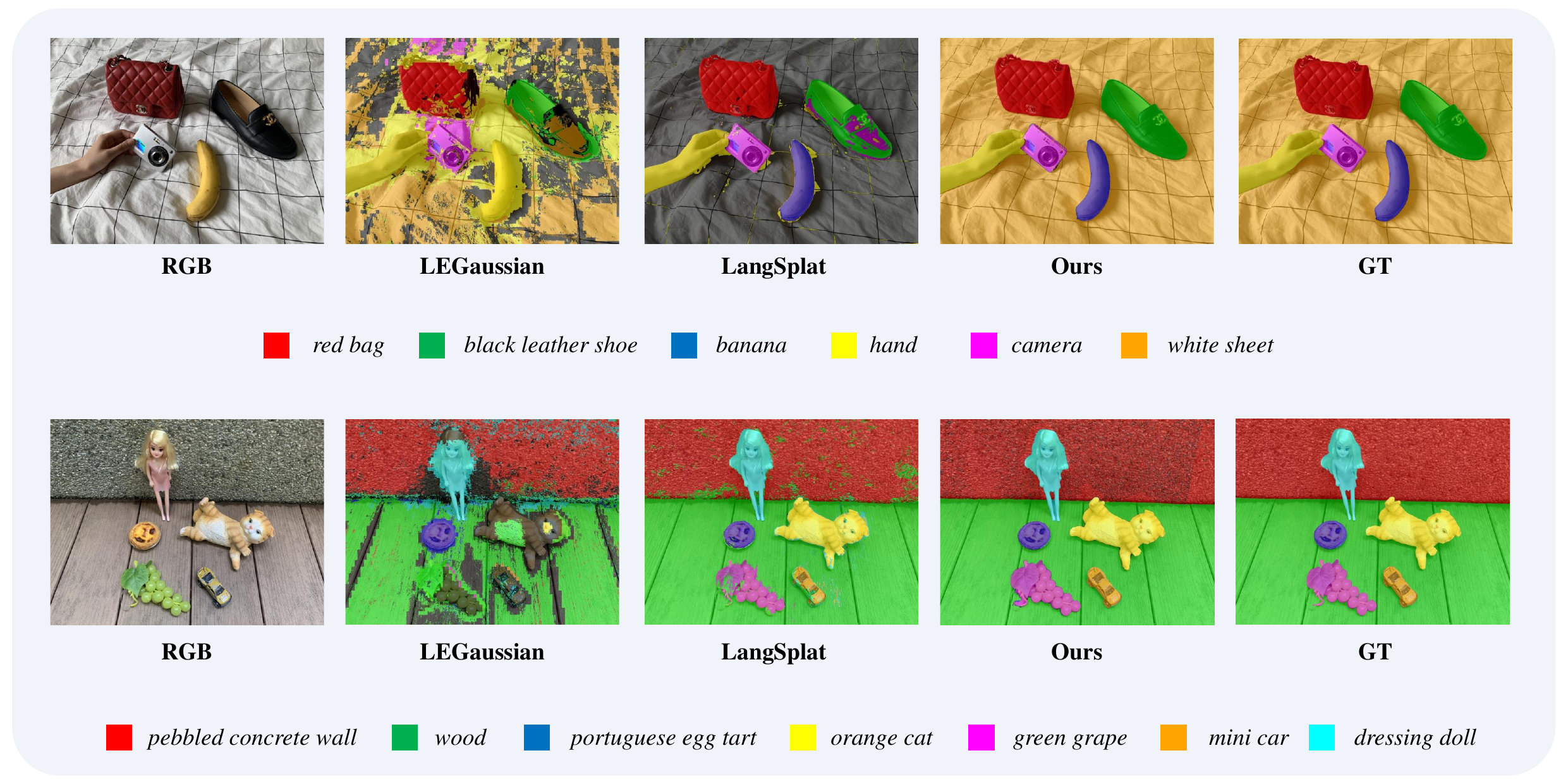}
    \vspace{-15pt}
    \caption{Open-vocabulary 3D semantic segmentation on the 3D-OVS dataset.}
    \label{fig:3dovs_ex}
\end{figure*}

\subsubsection{3D-OVS Dataset}
We also compare SparseLGS with other methods on the 3D-OVS dataset. Unlike the original 3D-OVS~\cite{3dovs} method, which requires obtaining a complete list of objects in the scene beforehand, we only use textual descriptions to query and obtain object masks for all methods.

Table~\ref{tab:3dovs} presents the numerical results of our method compared to other SOTA 3D language field reconstruction methods. We used only three views as input for experiments on the 3D-OVS dataset. It can be observed that our method has achieved great results and has strong numerical stability across different datasets. Other methods, such as LangSplat, rely on autoencoders to reconstruct the original CLIP features. As a result, their performance is directly influenced by the quality of the autoencoder, leading to fluctuating results. In the case of the `room' dataset, this dependence on the autoencoder caused complete prediction failures. Despite multiple repeated training experiments, we were unable to achieve satisfactory results. This further emphasizes the necessity of establishing a direct mapping from high-dimensional to low-dimensional spaces in our method.

Fig.~\ref{fig:3dovs_ex} and the blue box of Fig.~\ref{fig:3dovs_supp} showcase the superiority of our method in terms of visual results. It demonstrates better boundary information and regularization of masks.

\begin{table}[htp]
\centering
\caption{Ablation Study on different dimension reduction methods of CLIP features. Bij. refers to our method's bijection, AE stands for autoencoder, and the third line represents using the original CLIP feature for training without any dimensionality reduction.}
\resizebox{0.86\linewidth}{!}{
\begin{tabular}{c|cc|c}
    \multirow{2}{*}{Dataset} & \multicolumn{2}{c|}{LERF} & 3D-OVS\\
    & mIOU $\uparrow$ & mACC $\uparrow$ & mIOU $\uparrow$\\
    \hline
    w. Bij. & \textbf{0.538} & \textbf{0.786} & \textbf{0.844}\\
    w. AE & 0.463 & 0.640 & 0.655\\
    w.o. Dim re. & O.O.M. & O.O.M & O.O.M\\
\end{tabular}
}
\label{tab:ablation_bijection}
\end{table}

\subsubsection{MASt3R setting for other methods}
To better evaluate the contributions of modules beyond MASt3R initialization, we conducted more qualitative and quantitative comparisons using the results of other methods (LangSplat, LEGaussian) initialized with MASt3R (here `-s' means use MASt3R).
\begin{table}[htp]
\centering
\caption{Quantitative results about MASt3R with other methods.}
\resizebox{\linewidth}{!}{
\begin{tabular}{c|cc|cc|cc}
    \multirow{2}{*}{ } & \multicolumn{2}{c|}{Ours} & \multicolumn{2}{c|}{LangSplat-s} & \multicolumn{2}{c}{LEGaussian-s}\\
    & mIOU $\uparrow$ & mACC $\uparrow$ & mIOU $\uparrow$ & mACC $\uparrow$ & mIOU $\uparrow$ & mACC $\uparrow$\\
    \hline
    LERF Dataset & \textbf{0.538} & \textbf{0.770} & 0.426 & 0.636 & 0.237 & 0.632 \\
    3DOVS Dataset & \textbf{0.844} & - & 0.645 & - & 0.434 & - \\
\end{tabular}
}
\label{tab:rebuttal_mast3r}
\end{table}

\begin{figure*}[tb]
    \centering
    \includegraphics[width=\linewidth]{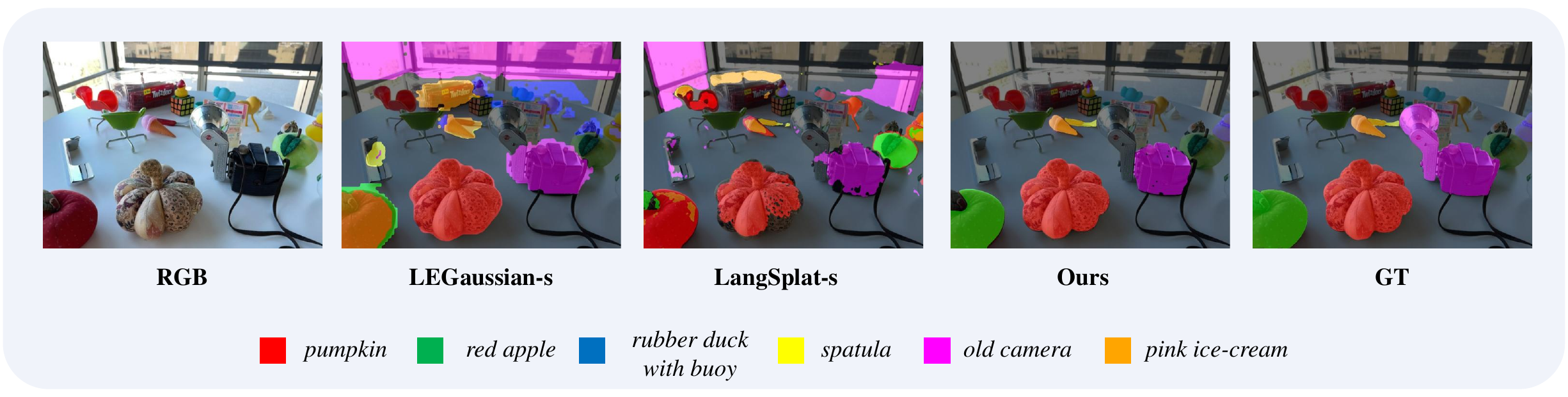}
    \vspace*{-17pt}
    \caption{Qualitative results about MASt3R with other methods.}
    \label{fig:fig_rebuttal_sparse}
\end{figure*}

Table~\ref{tab:rebuttal_mast3r} and Fig~\ref{fig:fig_rebuttal_sparse} show that adding MASt3R initialization to other methods does not fully close the semantic-quality gap, further highlighting the role of our semantic alignment and bijection modules.

\subsubsection{Initialization Model Ablation}
\revadd{
To evaluate the influence of geometry initialization, we replace the initializer while keeping the same four Teatime input views, semantic features, and downstream SparseLGS training pipeline. Table~\ref{tab:init_model_ablation} reports the full-evaluation results. Different initializers provide different camera parameters and point clouds, and each has its own strengths and errors. As a result, the final scores are close: MASt3R gives a slight numerical advantage, while DUSt3R and VGGT also remain effective. This indicates that geometry initialization affects the final semantic field, but the subsequent semantic alignment and feature representation can work with different learning-based geometry initializers.
}

\begin{table}[htp]
\centering
\caption{\revadd{Initialization model ablation on Teatime.}}
\resizebox{0.82\linewidth}{!}{
\revadd{
\begin{tabular}{c|ccc}
    Init. model & Init. points & mIOU $\uparrow$ & LocAcc $\uparrow$\\
    \hline
    DUSt3R & 536,922 & 0.6977 & 0.8684\\
    MASt3R & 495,189 & \textbf{0.7125} & 0.8684\\
    VGGT & 327,150 & 0.7063 & \textbf{0.8947}\\
\end{tabular}
}}
\label{tab:init_model_ablation}
\end{table}

\subsubsection{Comparison with Feed-forward LSM}
\revadd{
Table~\ref{tab:lsm_compare} summarizes our additional comparison with LSM~\cite{fan2024largespatialmodel}. The ScanNet part uses the same sparse-view split for all methods. LSM is a general feed-forward model trained to lift LSeg features, while SparseLGS performs per-scene optimization and can use different 2D semantic sources. Under the LSeg-based ScanNet protocol, SparseLGS(LSeg) is close to LSM but slightly lower. One reason is that region-level SAM+LSeg supervision is not always well aligned with dense ScanNet labels. To better understand this factor, we additionally test SparseLGS(GTseg) with ScanNet GT segmentation as an oracle semantic source. The improvement suggests that the per-scene Gaussian semantic field can benefit from more reliable 2D semantic assignment. On LERF Teatime, the scene distribution and fine-grained object prompts fall outside the fixed ScanNet label space, whereas SparseLGS builds scene-specific open-vocabulary features and remains effective.
}

\begin{table}[htp]
\centering
\caption{\revadd{Additional comparison with the feed-forward LSM baseline. Acc. denotes pixel accuracy for ScanNet and localization accuracy for LERF Teatime.}}
\resizebox{\linewidth}{!}{
\revadd{
\begin{tabular}{c|c|cc}
    Dataset & Method & mIOU $\uparrow$ & Acc. $\uparrow$\\
    \hline
    ScanNet & LSM & 0.4709 & 0.7391\\
    ScanNet & SparseLGS(LSeg) & 0.4527 & 0.7054\\
    ScanNet & SparseLGS(GTseg) & \textbf{0.5977} & \textbf{0.9055}\\
    \hline
    LERF Teatime & LSM & 0.0006 & 0.0922\\
    LERF Teatime & LSeg & 0.1447 & 0.3144\\
    LERF Teatime & SparseLGS & \textbf{0.7125} & \textbf{0.8684}\\
\end{tabular}
}}
\label{tab:lsm_compare}
\end{table}

\revadd{
The ScanNet comparison follows LSM's fixed eight-label semantic space, including \textit{wall}, \textit{floor}, \textit{ceiling}, \textit{chair}, \textit{table}, \textit{sofa}, \textit{bed}, and \textit{other}. LSM learns semantic Gaussians from LSeg feature supervision, whereas SparseLGS(LSeg) uses region-level SAM+LSeg features in a per-scene optimization pipeline. The slight gap between these two rows is therefore partly related to the 2D semantic source. In our diagnostic on LSM-cropped ScanNet frames, direct SAM+LSeg labels reach around 0.67 mIoU against ScanNet labels. This result is reasonable but still shows that SAM regions and LSeg predictions are not perfectly aligned with the dense ScanNet label protocol. To reduce the influence of segmentation-induced errors and better isolate the behavior of the 3D semantic field, we additionally evaluate SparseLGS(GTseg) with ScanNet GT segmentation as an oracle semantic source. The higher result indicates that SparseLGS benefits from more accurate 2D semantic assignment, and that part of the gap in SparseLGS(LSeg) comes from the semantic-source bottleneck rather than the per-scene field representation itself. This row should be interpreted as an oracle analysis rather than the default open-vocabulary setting. The LERF Teatime LSeg row directly applies LSeg to RGB images to generate dense feature maps and then performs the same object-query evaluation. These comparisons are intended to clarify that LSM targets feed-forward generalization in a fixed ScanNet/LSeg label space, while SparseLGS targets a per-scene optimized open-vocabulary language field.
}

\subsection{Ablation Study}
\paragraph{Ablation studies on Bijection.} Table~\ref{tab:ablation_bijection} presents the numerical results obtained using different dimensionality reduction methods. It can be observed that training without reducing the dimension of the feature will lead to `CUDA OUT OF MEMORY' (denoted by O.O.M). We can also know that constructing a low-dimensional to high-dimensional bijection significantly outperforms the use of an autoencoder for language field reconstruction.

\begin{figure}
    \centering
    \includegraphics[width=\linewidth]{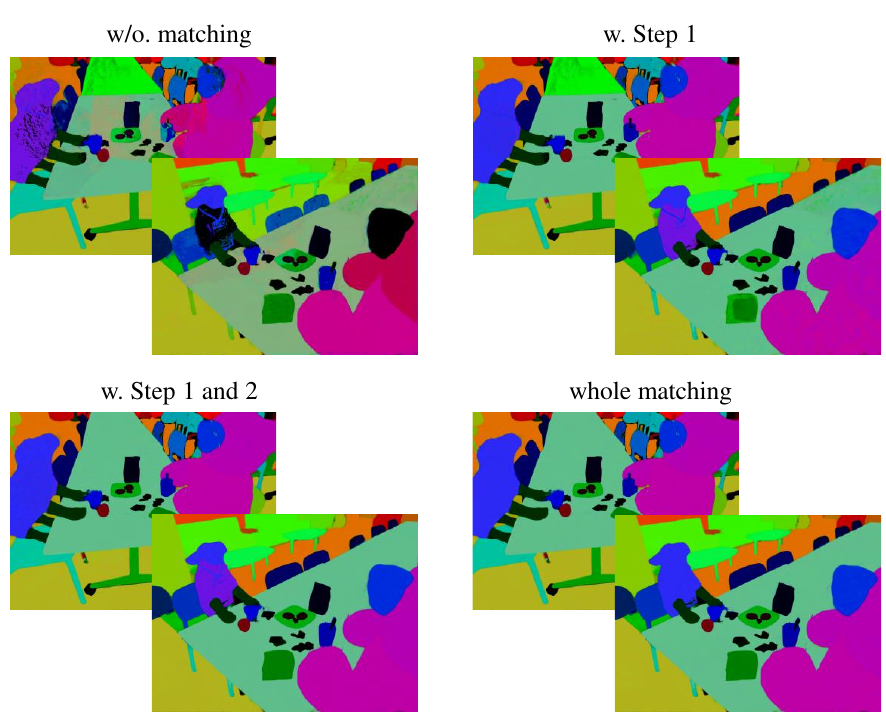}
    \caption{Ablation Study on Multi-view semantic alignment, where we visualize and compare the semantic results obtained at each of the three matching steps.}
    \label{fig:ablation_match}
\end{figure}

\begin{table}[htp]
\centering
\caption{Ablation on each step in multi-view semantic alignment. SIM. denotes semantic cosine similarity.}
\resizebox{\linewidth}{!}{
\begin{tabular}{c|ccccc}
    \multirow{2}{*}{ } & w/o. align & w. step1 & w. step1,2 & w. step3 & w. step1,2,3\\
    \hline
    mIOU $\uparrow$ & 0.496  & 0.511  & 0.515  & 0.507 & \textbf{0.538} \\
    mACC $\uparrow$ & 0.717  & 0.763  & 0.763  & 0.746 & \textbf{0.786} \\
    \hline \hline
    SIM. $\uparrow$ & 0.952  & 0.958  & 0.960  & 0.963 & \textbf{0.969} \\
    $\textrm{MSE}_{sem}$ $\downarrow$ & $8.59*10^{-3}$  & $6.05*10^{-3}$  & $5.60*10^{-3}$ & $6.02*10^{-3}$ & $\mathbf{4.99*10^{-3}}$ \\
\end{tabular}
}
\label{tab:ablation_alignment}
\end{table}

\paragraph{Ablation studies on Multi-view semantic alignment.}
We conduct ablations on the LERF Dataset and present the visualization results in Fig.~\ref{fig:ablation_match}. We conducted experiments without and with matching up to step 1-3, respectively. The results showed that step 1 significantly corrected many matching errors. Step 2 further merged regions with inconsistent SAM segmentation granularity, while step 3 fine-tuned the matching results on a spatial level. the quantitative results of Table~\ref{tab:ablation_alignment} demonstrate the contribution of each step. The first two rows show the improvements of consistency operations in downstream tasks, while the last two rows demonstrate the enhancement of consistency with the ground truth semantic map (semantic features have been normalized to [0, 1]). Notably, since the query targets in the dataset mainly focus on small objects, such as plates, bear noses, cookies, etc., the numerical results do not fully reflect the improvements in all regions.

\revadd{
\paragraph{Ablation on the number of input views.}
To analyze whether more input views improve the final semantic field, we conduct a controlled view-number ablation on the Teatime scene. As shown in Table~\ref{tab:view_number_ablation}, the semantic mIoU remains stable from two to six views. This suggests that SparseLGS can use additional sparse views without a clear degradation when the geometry complexity is controlled, but the extra views do not automatically translate into a large gain. In practice, more views provide more coverage but also introduce more SAM/CLIP inconsistencies and matching paths; therefore, the benefit depends on whether the added observations are reliable.
}

\begin{table}[htp]
\centering
\caption{\revadd{Controlled view-number ablation on Teatime.}}
\resizebox{0.70\linewidth}{!}{
\revadd{
\begin{tabular}{c|cc}
    Input views & mIOU $\uparrow$ & LocAcc $\uparrow$\\
    \hline
    2 & 0.7341 & 0.9000\\
    4 & 0.7264 & 0.9000\\
    6 & 0.7313 & 0.8500\\
\end{tabular}
}}
\label{tab:view_number_ablation}
\end{table}

\revadd{
For a more controlled comparison, we report the results on the evaluation subset shared by the 2-, 4-, and 6-view settings, using the common 2-view evaluation frames and prompts. Increasing the number of views indeed brings more geometric and semantic constraints, including more observed surfaces and more cross-view evidence. At the same time, it can also introduce additional semantic-training noise, such as local floaters from feed-forward geometry initialization, more possible matching paths, and more inconsistent SAM/CLIP region features. In this experiment, these positive and negative factors largely offset each other, leading to similar final scores across the tested view counts.

We further tested point-cloud capping on the six-view setting. Mildly reducing the initialized points from uncapped to 500K improved the full-evaluation mIoU from 0.5967 to 0.6020, while overly aggressive capping degraded the result to 0.5358 at 50K points. This indicates that the initialization has a useful operating range: too many Gaussians may absorb noisy semantic correspondences, but too few Gaussians enlarge splats and blur fine semantic boundaries.
}

\begin{figure}
    \centering
    \includegraphics[width=\linewidth]{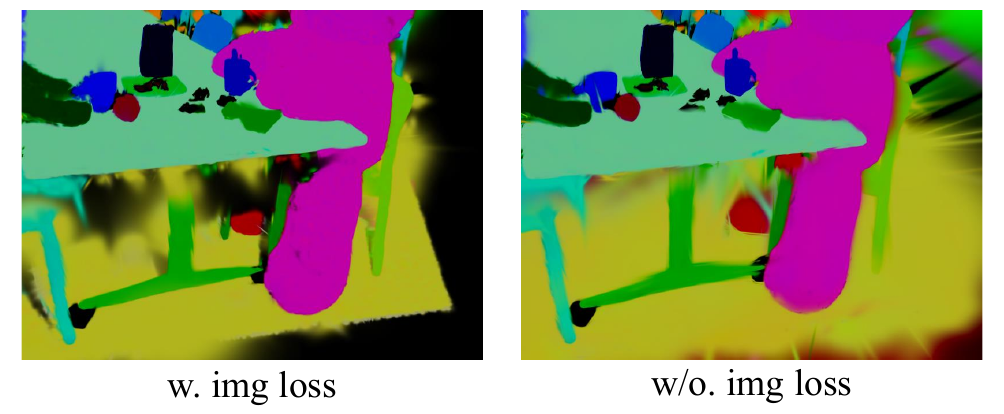}
    \caption{Ablation Study on the effect of image loss during language field training.}
    \label{fig:ablation_train}
\end{figure}

\paragraph{Ablation studies on language field training.}
We compare the results of training language field with image loss and training singly with semantic loss for novel view synthesis. From Fig.~\ref{fig:ablation_train}, it can be observed that the stronger geometric constraint brought by the image loss prevents the Gaussian shape from growing uncontrollably and overfitting, which makes the Gaussian distributions in space more reasonable in terms of shape and positional distribution. Without image loss (right side of Fig.~\ref{fig:ablation_train}), Gaussians are freely added in unseen regions (black areas), which has limited impact on numerical performance of training views but degrades novel view synthesis. This is clearly shown in the supplementary video, where our semantic field yields cleaner edges and avoids misplaced Gaussians in occluded or invisible regions.

\section{Discussion}
\subsection{Difference From Baseline}
\begin{figure}[htp]
    \centering
    \includegraphics[width=\linewidth]{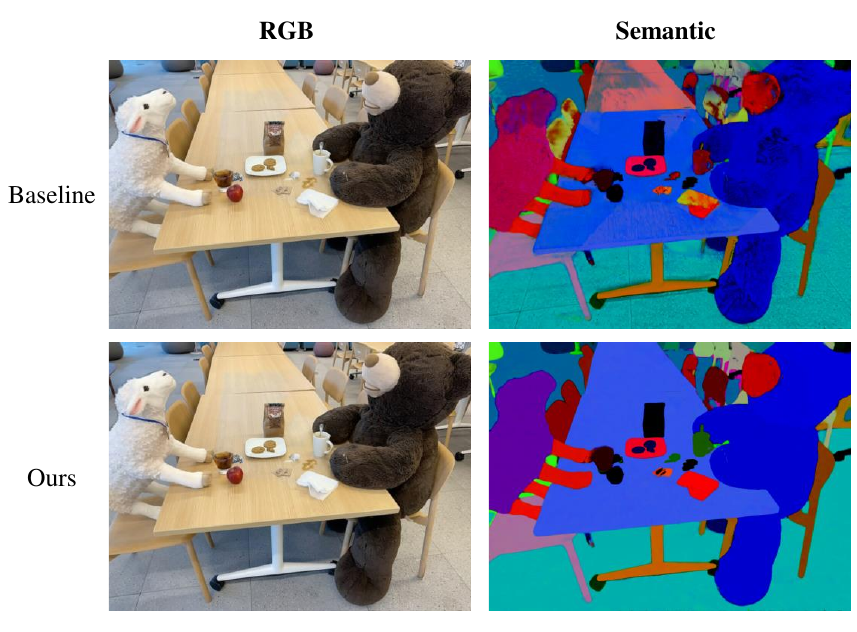}
    \caption{Comparison with baseline on RGB and semantic. The contribution of our work focuses on semantic improvement.}
    \label{fig:rgb_sem}
\end{figure}
We use MASt3R + LangSplat as our baseline. A simple combination of these two methods shows some improvement over not using MASt3R, but it still falls short compared to our whole pipeline. The reason lies in our choice of bijection and multi-view semantic alignment for addressing the sparsity issue. Tab. \ref{tab:ablation_bijection},\ref{tab:ablation_alignment} show the quantitative improvements from our proposed modules. The contribution of our work focuses on the semantic part, while the RGB part follows the baseline's performance. As shown in Fig.~\ref{fig:rgb_sem}, our method has significant qualitative improvements in semantic results, with many regions (such as tables, sheep) being more consistent. Since the query targets in the dataset are mainly small objects (plates, bear noses, cookies, etc.), the numerical results do not fully reflect the improvements in larger regions.

\subsection{About Learning Based Stereo Model}
We adopt a learning-based stereo model instead of COLMAP because SFM+MVS lack the capability to estimate camera poses and point clouds from sparse view inputs. In our framework, we use MASt3R~\cite{mast3r_arxiv24} to get these information and further optimize them in our initial Gaussians training step. DUSt3R~\cite{dust3r_cvpr24, dust3r_arxiv23}, VGGsfM~\cite{wang2024vggsfm}, VGGT~\cite{wang2025vggt} or other similar methods can also serve as viable alternatives.
\revadd{The stereo model is used to initialize camera poses and point clouds. As shown in Table~\ref{tab:init_model_ablation}, replacing the MASt3R initializer with DUSt3R or VGGT yields close semantic results, although each initializer produces different camera and point-cloud errors. This suggests that our semantic alignment and feature representation can work with different learning-based geometry initializers. More broadly, each upstream component in our pipeline can be replaced by a stronger alternative, including the 2D segmentation model, the semantic feature extractor, and the camera/point-cloud initializer.}

\subsection{More Explanation about semantic alignment}
We achieve 3D semantic consistency through a soft constraint based on multi-view projection. Multi-view consistency at the 2D level critically affects the distillation of planar semantic information into 3D. Semantic misalignment of the baseline method, as seen in the upper-left of Fig.~\ref{fig:ablation_match}, differ noticeably from the training semantic map. This misalignment also affects the quality of downstream task results (left, right column of Table~\ref{tab:ablation_alignment}). By leveraging soft constraint based on multi-view projection to optimize the semantic field, we can achieve higher reconstruction quality.

\subsection{Failure Cases and Artifacts}
\begin{figure}[H]
    \centering
    \includegraphics[width=0.96\linewidth]{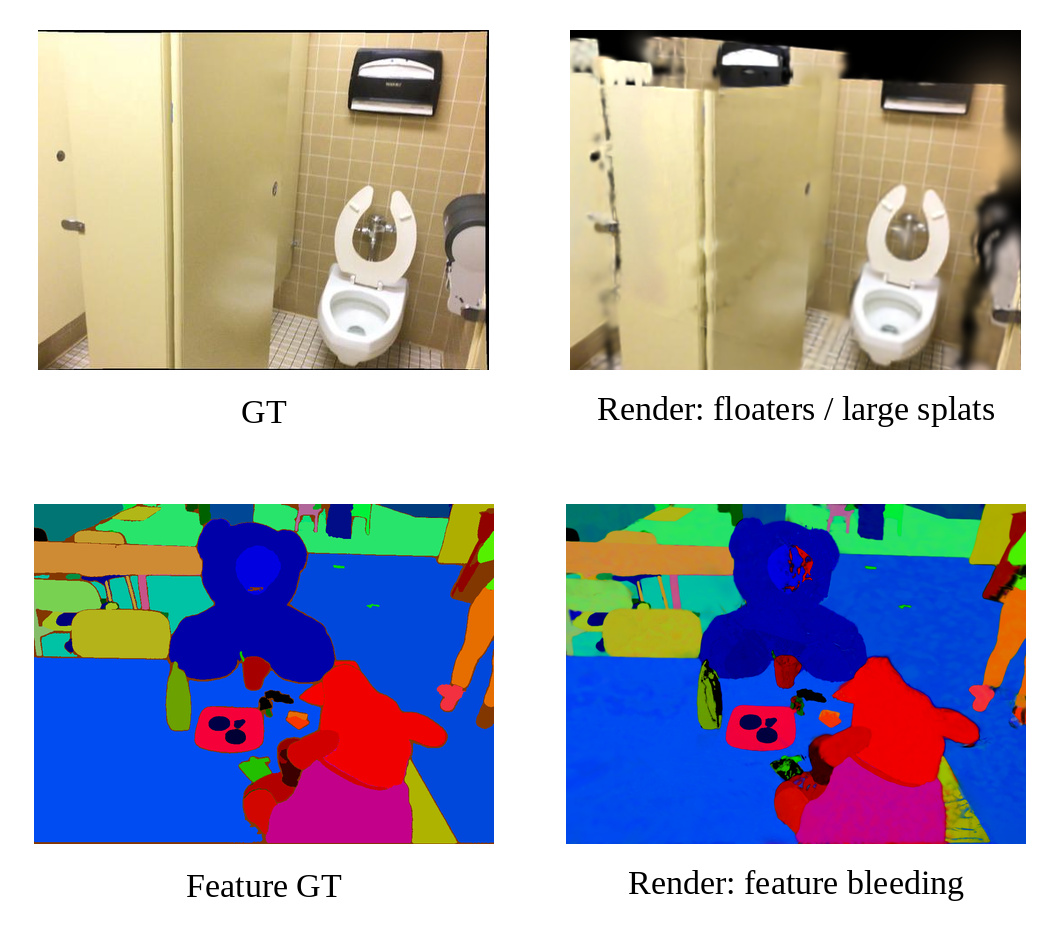}
    \caption{\revadd{Representative failure cases and artifacts. Sparse-view observations may leave some novel-view regions weakly constrained, producing floaters or enlarged splats. Semantic feature rendering can also show boundary bleeding when sparse-view region features are inconsistent.}}
    \label{fig:failure_artifacts}
\end{figure}
\revadd{Fig.~\ref{fig:failure_artifacts} shows representative limitations. SparseLGS generally reconstructs coherent semantic fields from a few pose-free images, but it can still fail when the target region is severely occluded, only partially observed, or surrounded by visually and semantically similar objects. In novel views that reveal weakly constrained regions, some Gaussians may become overly large or misplaced, causing floaters or smeared surfaces. Thin structures and small objects are also challenging because SAM masks and CLIP region features can be inconsistent across viewpoints. These artifacts indicate that sparse-view semantic field reconstruction is jointly limited by geometric initialization, view coverage, and 2D semantic ambiguity.}

\section{Conclusion}
We proposed SparseLGS, a method for sparse, pose-free view 3D language field reconstruction that supports open-vocabulary queries. We combined multi-view stereo models to obtain a strong prior from sparse and pose-free inputs and then addressed the challenge of multi-view inconsistency caused by CLIP and SAM via multi-view semantic alignment. By establishing a bijective mapping, we transformed the high-dimensional CLIP features into low-dimensional features that served as Gaussian attributes for tile-based rendering. We combined image loss with semantic loss to produce a high-quality 3D language field. Compared to existing methods, our method requires only a small number of input views (3-4 views) to support highly accurate open vocabulary queries and is at least 5 times faster than existing SOTA methods.

\ifCLASSOPTIONcompsoc
  \section*{Acknowledgments}
\else
  \section*{Acknowledgment}
\fi
This research was supported by the National Natural Science Foundation of China (No.62272433), Anhui Provincial Natural Science Foundation (No.2508085ZD011), Key Science \& Technology Project of Anhui Province (No. 202523o09050004) and the Fundamental Research Funds for the Central Universities.

{
\bibliographystyle{IEEEtran}
\bibliography{main}
}

\newpage

\begin{IEEEbiography}[{\includegraphics[width=0.82in,height=1.03in,clip,keepaspectratio]{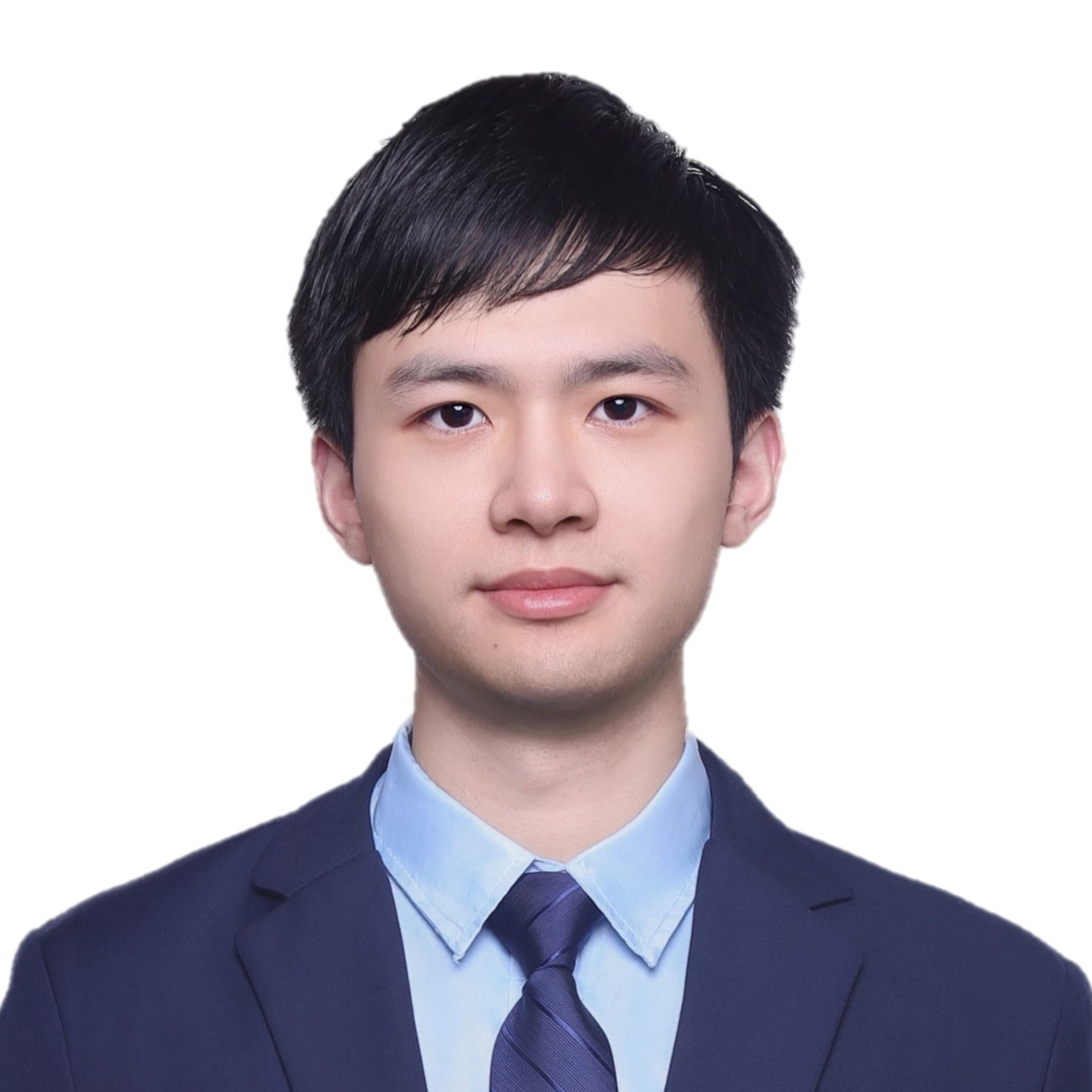}}]{Jun Hu} is a master student at the School of Mathematica Sciences, University of Science and Technology of China. His research interests include multi-view human synthesis and 3D reconstruction in computer vision.
\end{IEEEbiography}\vspace{-8pt}

\begin{IEEEbiography}[{\includegraphics[width=0.82in,height=1.03in,clip,keepaspectratio]{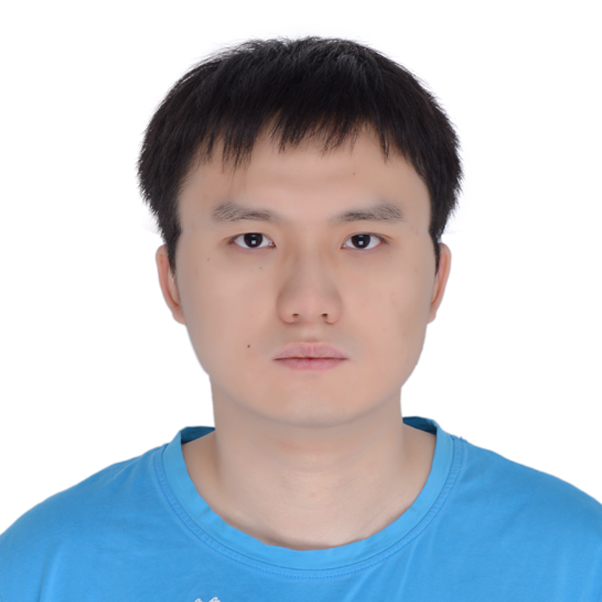}}]{Zhang Chen} is a research engineer at Meta. He received his Ph.D. from ShanghaiTech University in 2022. Before joining Meta, he worked as a senior research engineer at OPPO US Research Center. His research interests include computer vision, computer graphics, and computational photography, with a focus on neural rendering and generation.
\end{IEEEbiography}\vspace{-8pt}

\begin{IEEEbiography}[{\includegraphics[width=0.82in,height=1.03in,clip,keepaspectratio]{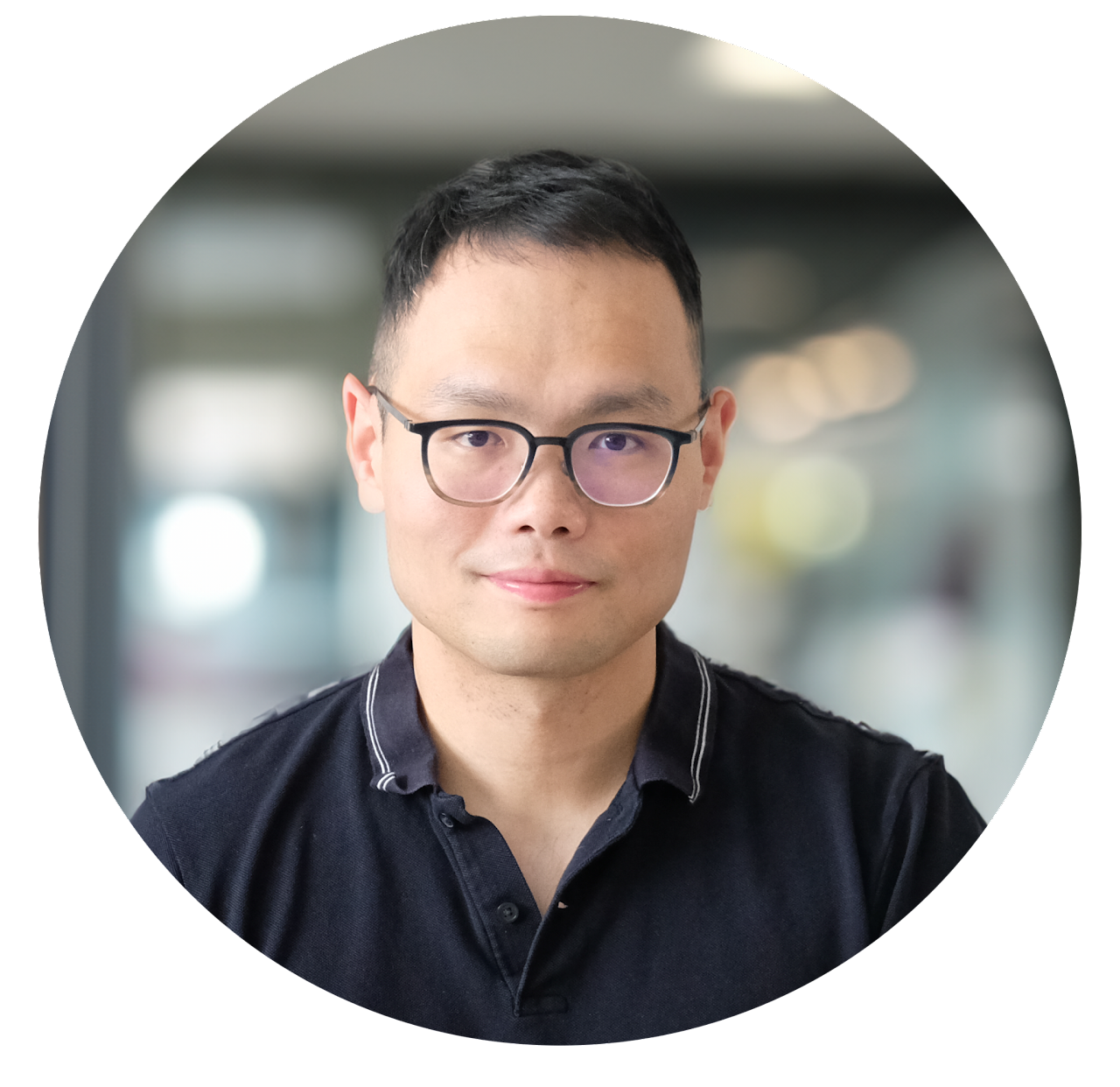}}]{Zhong Li}
is a Senior Research Engineer at Apple Inc., specializing in Generative AI. He holds a Ph.D. from the University of Delaware and previously worked as a Senior Staff Research Scientist at OPPO US Research Center, where he contributed to advancing mobile AR/VR technologies. His research interests include computer graphics, 3D vision, and computational imaging, with expertise in neural rendering, 3D generative AI, image-based rendering, relighting, and modeling 2D/3D human faces and bodies. His work has been published in top venues such as TPAMI, TVCG, SIGGRAPH, CVPR, ICCV, ECCV, and NeurIPS, and he has served as a reviewer/program committee for these conferences. He received the Best Paper Award at ACM VRCAI in 2019.
\end{IEEEbiography}\vspace{-8pt}

\begin{IEEEbiography}[{\includegraphics[width=0.82in,height=1.03in,clip,keepaspectratio]{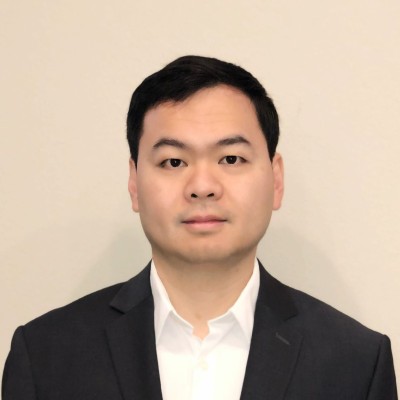}}]{Yi Xu}
is the VP of Alpha Labs at Goertek. His research interest lies in 3D computer graphics and computer vision, with a focus on Extended Reality. Before joining Goertek, Dr. Xu worked at various industrial labs such as GE Research, JD.COM Silicon Valley Labs, and OPPO US Research Center. Dr. Xu got his Ph.D. degree from Purdue University in 2010.
\end{IEEEbiography}\vspace{-8pt}

\begin{IEEEbiography}[{\includegraphics[width=0.82in,height=1.03in,clip,keepaspectratio]{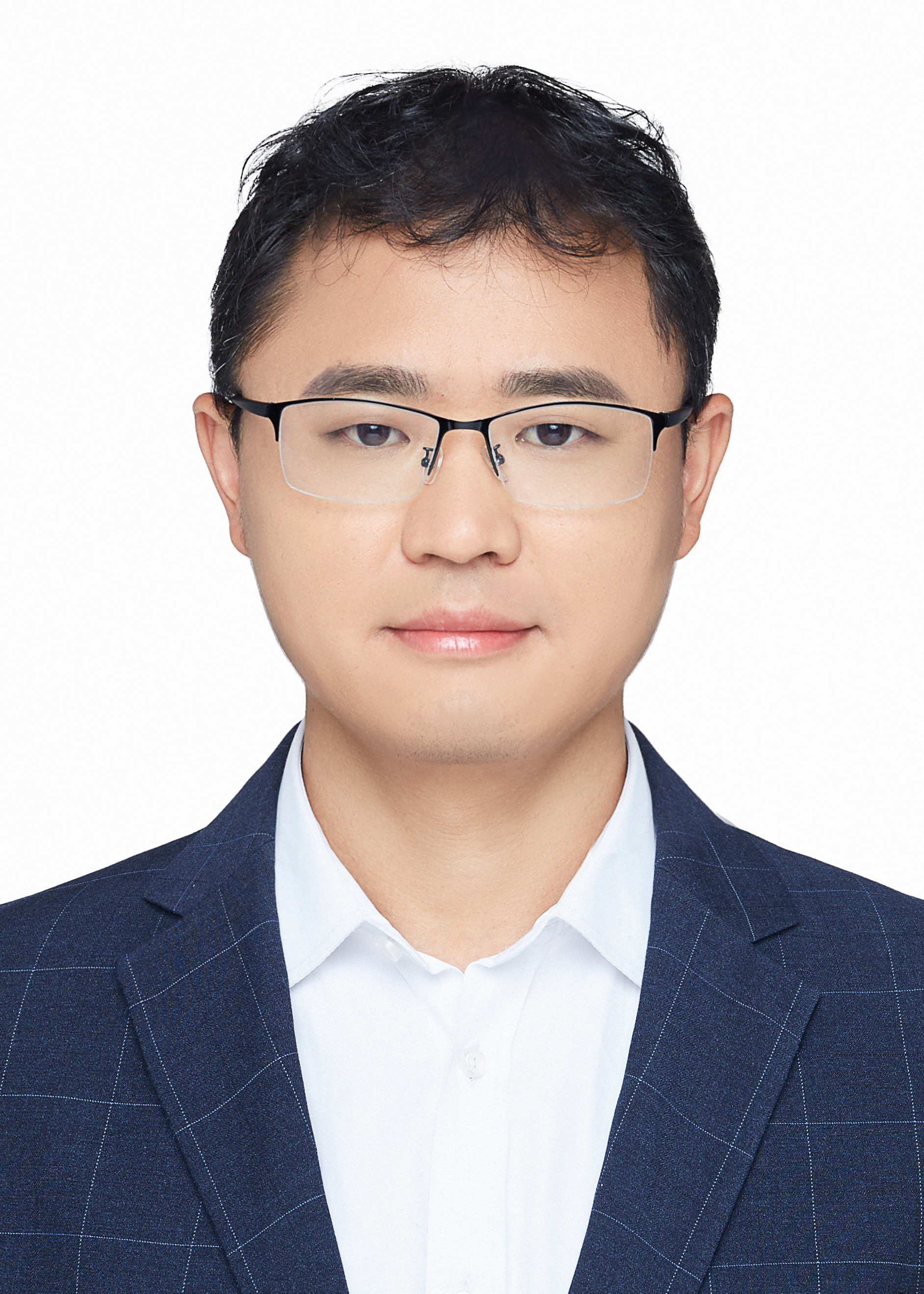}}]
{Juyong Zhang}\looseness=-1 is a professor in the School of Mathematical Sciences at University of Science and Technology of China. He received the BS degree from the University of Science and Technology of China in 2006, and the Ph.D degree from Nanyang Technological University, Singapore. He mainly conducts research at the intersection of Vision, Graphics, and AI with a special focus on capturing, modeling and synthesizing objects, humans and large-scale scenes. He is an associate editor of IEEE Transactions on Mobile Computing and IEEE Computer Graphics and Applications.
\end{IEEEbiography}

\vfill

\end{document}